\documentclass[letterpaper, 10 pt, conference]{ieeeconf}

\IEEEoverridecommandlockouts                              

\usepackage{cite}
\usepackage{float}
\usepackage{epsfig} 
\usepackage{graphics} 
\usepackage[utf8]{inputenc} 
\usepackage{amsmath,amssymb,amsfonts}
\usepackage{comment}
\usepackage{siunitx}
\DeclareSIUnit{\rad}{rad}
\usepackage{bm}
\usepackage[dvipsnames]{xcolor}

\newcommand{\momenta}{p}
\newcommand{\JointHamiltonian}{\mathcal{H}_q}
\newcommand{\ImpedanceHamiltonian}{\mathcal{H}_Z}
\newcommand{\CausalHamiltonian}{\mathcal{H}_\Omega}
\newcommand{\CausalDMomenta}{\bm{\momenta} - \bm{\momenta}_d}
\newcommand{\CausalDeviation}{\bm{x} - \bm{x}_d}

\title{\LARGE \bf
Leveraging Port-Hamiltonian Theory for Impedance Control Benchmarking
}

\author {
    Leonardo F. Dos Santos, Elisa G. Vergamini, Cícero Zanette, Lucca Maitan, and Thiago~Boaventura
    \thanks{This work was partially supported by São Paulo Research Foundation (FAPESP) under grants 2018/15472-9, 2021/03373-9, 2021/09244-6 and 2023/11407-6, by Brazilian Federal Agency for Support and Evaluation of Graduate Education (CAPES) under grant 88887.817139/2023-00, and by Petrobras/ANP under grant 2023/00494-5}
    \thanks{Authors are with the São Carlos School of Engineering, University of São Paulo, Brazil. Email: {\tt\small leonardo.felipe.santos@usp.br}}
}

\begin{document}

\maketitle
\thispagestyle{empty}
\pagestyle{empty}

\begin{abstract}


This work proposes PH-based metrics for benchmarking impedance control. A causality-consistent PH model is introduced for mass-spring-damper impedance in Cartesian space. Based on this model, a differentiable, force-torque sensing-independent, n-DoF passivity condition is derived, valid for time-varying references. An impedance fidelity metric is also defined from step-response power in free motion, capturing dynamic decoupling. The proposed metrics are validated in Gazebo simulations with a six-DoF manipulator and a quadruped leg. Results demonstrate the suitability of the PH framework for standardized impedance control benchmarking.

\end{abstract}

\section{Introduction}


Impedance control is a versatile method for shaping the dynamic behavior of legged robots, manipulators, and haptic devices in unknown environments ranging from space to underwater applications \cite{risiglione2022whole, zhang2023model, pedro2024quadruped}. However, this versatility poses a challenge: developing a uniform assessment of performance and stability that is suitable for such a broad range of applications. This challenge motivates the standardization of benchmarks, or test procedures with evaluation metrics that enable comparison and reproducibility in research.


Benchmarking efforts in manipulation with industrial robots illustrate this trade-off between application-specific and general applicability. For example, \cite{behrens2018performance} proposed a repeatability metric for force-controlled contour-following, while \cite{finkbeiner2023concept} defined procedures aligned with ISO 9283 for trajectory tracking. NIST, in turn, introduced a benchmark for evaluating end-effector grasping \cite{falco2020benchmarking}.


On the other hand, benchmarking methodologies in the research landscape align more closely with the rationale of impedance control. Implementing this method requires the actuator dynamics at the joint level to be negligible; however, this assumption is difficult to realize in hydraulic or cable-driven actuators. To address this issue, \cite{vergamini2024force} proposed a force control benchmark. Complementarily, \cite{suarez2020benchmarks} introduced benchmarks for aerial manipulation with an impedance-controlled arm on a drone, addressing task-level dynamics.


Although these benchmarks considered physical aspects such as end-effector position and force, and engineering aspects such as repeatability, the central aspect of impedance control is the relationship between position error and interaction force \cite{hogan1985PartI, hogan1987modularity}. This shifts the problem toward system identification rather than the assessment of unrelated variables. Since this relationship is defined by a mechanical model, physics-oriented control formulations, such as the Port-Hamiltonian (PH) framework, are well suited to describe and characterize it.


The PH formalism extends classical Hamiltonian systems by incorporating an input-output space and supporting nonlinear state-space systems. Hamiltonian systems already play a well-established role in \textit{physics-informed machine learning}, embedding prior knowledge from first principles into learning algorithms \cite{chen2021data, celledoni2023learning}, thereby connecting nearly two centuries of classical mechanics with modern machine learning (ML). PH-based machine learning has also gained attention: \cite{beckers2022gaussian} proposed a Bayesian learning scheme for Gaussian process PH systems, while \cite{ortega2024learnability} analyzed the learnability of input-output dynamics for a linear PH system. In the model-based control domain, PH formulations support tuning rules for passivity-based control \cite{chan2023tuning} and the design of energy-tank-based impedance controllers \cite{mujica2021impedance, rashad2022energy}.


The PH framework relies on energy flow and incorporates the definition of passivity. It is therefore suitable for analyzing Z-width, i.e., the range of stiffness and damping that can be passively achieved. Z-width was originally introduced using linear control theory \cite{colgate1994coupled} and remains a standard tool for assessing the stability of impedance control. In \cite{hejrati2023nonlinear}, an experimental demonstration of the Z-width for a 7-DoF haptic impedance-controlled device. However, the Z-width criteria neglected the power injected by the controller reference. Herein, the passivity criteria include it. Evaluate passivity beyond linear theory and single-input/output domain is the way for broad passivity metrics. To the best of current knowledge, this has not yet been proposed. Given the many impedance control variants and applications, this work adopts a physics-centered, ML-friendly description—the PH framework—to derive and propose application-agnostic metrics for impedance control benchmarking. The contributions are:

\begin{itemize}
    \item Introducing a causality-consistent PH model for mass-spring-damper (MSD) impedance at Cartesian coordinates, with end-effector pose and momentum as inputs;
    \item Proposing a differentiable n-DoF passivity condition based on the causal PH model and robot mechanical energy, which accounts for the input energy from a time-varying reference and remains independent of force-torque sensing, even under Cartesian inertia shaping;
    \item Defining an impedance fidelity metric based on step-response power, comparing the power for a single Cartesian step against the full n-DoF impedance power.
\end{itemize}

Passivity and fidelity are demonstrated through simulations of a six-DoF manipulator and a quadruped robot leg.

\textit{Notation:} The $n \times n$ identity matrix is $\bm{I}_n$ and a $n \times n$ zero matrix is $\bm{0}_{n \times n}$. For $f : \mathbb{R}^n \rightarrow \mathbb{R} \hspace{1pt}$, $\nabla_{x}\,f := \partial f/ \partial x$, in column vector form. For $x \in \mathbb{R}^n$, $A$ is definite (semidefinite), $A \succ 0$ ($A \succeq 0$), if $A = A^T$ and $x^T A x > 0$ ($x^T A x \geq 0$) for all $x \in \mathbb{R}^n$ - $\{0_n\}$ ($\mathbb{R}^n$). 

\section{Theoretical Foundation}

Either Lagrangian or Hamiltonian differential equations describe analytical mechanics. The former set the well-known Euler-Lagrange equation:

\begin{equation}
    \bm{M}(\bm{q})\,\ddot{\bm{q}} + \bm{C}(\bm{q}, \dot{\bm{q}})\,\dot{\bm{q}} + \bm{g}(\bm{q}) = \bm{\tau} \label{eq:EulerLagrange}
\end{equation}

\noindent where $\bm{q} \in \mathbb{R}^n$ is the generalized configuration vector for \textit{n} degrees of freedom, $\bm{M}(\bm{q}) \succ 0 \hspace{1pt}, \in \mathbb{R}^{n \times n}$ is the generalized mass, $\bm{C}(\bm{q}, \dot{\bm{q}}) \in \mathbb{R}^{n \times n}$ is the Coriolis/centrifugal matrix, $\bm{g}(\bm{q}) \in \mathbb{R}^n$ is the vector of gravity forces, and $\bm{\tau} \in \mathbb{R}^n$ is the vector of generalized forces acting on the system. This approach sets \textit{n} 2\textsuperscript{nd} order differential equations representing a mechanical system, such as an \textit{n}-DoF robot.

\subsection{Robot Hamiltonian}

According to the Hamiltonian equations of motion \cite{goldstein1980classical, van2014port}, the state vector is $\begin{bmatrix}\bm{q} & \bm{\momenta}\end{bmatrix}^T$, where $\bm{\momenta} \in \mathbb{R}^n$ is the generalized momenta $\bm{\momenta} = \bm{M}(\bm{q}) \dot{\bm{q}}$. Then, \eqref{eq:EulerLagrange} is transformed into 2\textit{n} first order differential equations:

\begin{equation}
    \begin{bmatrix}
        \dot{\bm{q}} \\ \dot{\bm{\momenta}}
    \end{bmatrix}
    =
    \begin{bmatrix}
        \bm{0}_{n \times n} & \bm{I}_n \\
        -\bm{I}_n & -\bm{C}(\bm{q}, \dot{\bm{q}})
    \end{bmatrix}
    \begin{bmatrix}
        \nabla_{\bm{q}}\,\JointHamiltonian \\
        \nabla_{\bm{\momenta}}\,\JointHamiltonian
    \end{bmatrix}
    +
    \begin{bmatrix}
        \bm{0}_{n \times n} \\
        \bm{I}_n
    \end{bmatrix}
    \bm{\tau} \hspace{2pt},
    \label{eq:CanonicalHamiltonian}
\end{equation}

\noindent where $\JointHamiltonian: \mathbb{R}^n \times \mathbb{R}^n \rightarrow \mathbb{R}$ is the Hamiltonian of the system defined as the sum of the kinetic and potential (gravitational) energies:

\begin{equation}
    \JointHamiltonian(\bm{q}, \bm{\momenta}) = \frac{1}{2} \bm{\momenta}^T \bm{M}^{-1}(\bm{q}) \bm{\momenta} + U_g(\bm{q}) \hspace{2pt},
    \label{eq:nDoFHam}
\end{equation}

\noindent and $\nabla_{\bm{q}} \JointHamiltonian = \bm{g}(\bm{q})$, $\nabla_{\bm{\momenta}} \JointHamiltonian = \bm{M}^{-1}(\bm{q}) \bm{\momenta}$. The state space of \eqref{eq:CanonicalHamiltonian} is called the \textit{phase space}. Herein, the subscript $q$ is used in $\JointHamiltonian$ since it represents the mechanical energy in the joint space of the robot.

\subsection{Cartesian Impedance Hamiltonian}

 At the interaction port (end-effector), the desired model for impedance is the MSD in Cartesian coordinates:

\begin{align}
    \bm{f_{int}} &= \bm{\Lambda_d}\,\ddot{\bm{e}} + \bm{D_d}\,\dot{\bm{e}} + \bm{K_d}\,\bm{e} \label{eq:desired_dynamics} \\
    \bm{e} &= \bm{x} - \bm{x}_{d} \label{eq:error_definition}
\end{align}

\noindent where $\bm{e}$, and $\bm{f_{int}} \in \mathbb{R}^{k}$ are the deviation and the interaction vectors, respectively, and $\bm{\Lambda_d}$, $\bm{D_d}$, $\bm{K_d} \in \mathbb{R}^{k \times k}$ are $\succ 0$ matrices for the desired generalized mass, damping, and stiffness, respectively. In case of the \textit{k} dimensions include both translation and rotation, $\dot{\bm{x}}$ and $\dot{\bm{e}}$ are twists, while $\bm{f_{int}}$ is a wrench in the Cartesian space. The equilibrium point $\bm{x}_{d}$ defines where the end-effector is in the absence of an interaction wrench. Equation \eqref{eq:desired_dynamics} in Hamiltonian description is:

\begin{equation}
    \begin{bmatrix}
        \dot{\bm{e}} \\ \dot{\bm{\momenta}_e}
    \end{bmatrix}
    =
    \begin{bmatrix}
        \bm{0}_{k \times k} & \bm{I}_k \\
        -\bm{I}_k & -\bm{D_d}
    \end{bmatrix}
    \begin{bmatrix}
        \nabla_{\bm{e}}\,\ImpedanceHamiltonian \\
        \nabla_{\bm{\momenta}_e}\,\ImpedanceHamiltonian
    \end{bmatrix}
    +
    \begin{bmatrix}
        \bm{0}_{k \times k} \\
        \bm{I}_k
    \end{bmatrix}
    \bm{f_{int}} \hspace{2pt},
    \label{eq:CanonicalImpedance}
\end{equation}

\noindent with

\begin{equation}
    \ImpedanceHamiltonian = \frac{1}{2} \bm{\momenta}_e^T \bm{\Lambda_d}^{-1}\bm{\momenta}_e + \frac{1}{2} \bm{e}^T \bm{K_d}\bm{e} \hspace{2pt},
    \label{eq:ImpedanceHam}
\end{equation}

\noindent and $\bm{\momenta}_e = \bm{\Lambda_d} \dot{\bm{e}} \in \mathbb{R}^{k}$ is the error conjugate momenta. Unlike impedance causality, in this canonical form, the error is part of the state, while the interaction is part of the input. To preserve causality, at least the reference state must be included in the input vector, closely resembling the momentum transformation in \cite{ferguson2024port} for velocity-controlled actuators. However, this transformation does not apply here, since the end-effector state is still free to evolve, so $\bm{\momenta}_e$ is only partially defined by the input.

\subsection{Classical impedance control law}

Render \eqref{eq:desired_dynamics}, decoupled from the robot dynamics \eqref{eq:EulerLagrange}, is achieved by setting the joint torques as

\begin{multline}
    \bm{\tau}_{act} = \bm{g}(\bm{q}) +
    \bm{J}(\bm{q})^{T}\,[ \bm{\Lambda}(\bm{x})\,\ddot{\bm{x}}_{d} + \bm{\Gamma}(\bm{x}, \dot{\bm{x}})\,\dot{\bm{x}}\,- \\
    \bm{\Lambda}(\bm{x})\,\bm{\Lambda_d}^{-1}\,(\bm{D_d}\,\dot{\bm{e}} + \bm{K_d}\,\bm{e}) +
    (\bm{\Lambda}(\bm{x})\,\bm{\Lambda_d}^{-1} - \bm{I}_k)\,\bm{f_{int}} ]
    \label{eq:cartesian_impedance_law}
\end{multline}

\noindent where $\bm{\Lambda}(\bm{x})$ and $\bm{\Gamma}(\bm{x}, \dot{\bm{x}})$ are the robot’s generalized mass and Coriolis matrices in Cartesian coordinates, respectively, and $\bm{J}(\bm{q}) \in \mathbb{R}^{k \times n}$ is the geometric Jacobian. To avoid inertia shaping (IS) and feedback from the interaction wrench sensor, one can use

\begin{equation}
    \bm{\tau}_{act} = \bm{g}(\bm{q}) +
    \bm{J}(\bm{q})^{T}\big[\bm{\Lambda}\,\ddot{\bm{x}}_{d} + \bm{\Gamma}\,\dot{\bm{x}}\,- \bm{D_d}\,\dot{\bm{e}} - \bm{K_d}\,\bm{e} \big]
    \label{eq:cartesian_impedance_law_no_shaping}
\end{equation}

Equation \eqref{eq:cartesian_impedance_law} corresponds to the classical impedance control law, while \eqref{eq:cartesian_impedance_law_no_shaping} represents the variant without IS. See \cite{ott2008cartesian} for further details.

\subsection{Port-Hamiltonian Description}

An explicit PH system is an input-state-output continuous time system defined by \cite{secchi2007control}:

\begin{align}
    \mathbf{\dot{x}} &= \left[ \mathbf{C(x)} - \mathbf{R(x)} \right] \nabla_{\mathbf{x}} \mathcal{H} + \mathbf{G(x)} \mathbf{u} \\
    \mathbf{y} &= \mathbf{G}^T(\mathbf{x}) \nabla_{\mathbf{x}} \mathcal{H}
\end{align}

\noindent where $\mathbf{x}$, $\mathbf{u}$, and $\mathbf{y}$ are the state, the input, and the output vectors, respectively. The matrix $\mathbf{C(x)} = - \mathbf{C^T(x)}$ describes the non-dissipative connections, $\mathbf{R(x)} \succeq 0$ represents the dissipative connections, and $\mathbf{G(x)}$ is the matrix of external inputs. The power exchange $\mathbf{y}^T \mathbf{u}$, the system energy, and the dissipation are related by:

\begin{equation}
    \mathbf{y}^T \mathbf{u} = \frac{d \mathcal{H}}{dt} + \nabla^T_{\mathbf{x}} \mathcal{H}\,\mathbf{R(x)}\,\nabla_{\mathbf{x}} \mathcal{H}
    \label{eq:EnergyBalance}
\end{equation}

The rightmost term in \eqref{eq:EnergyBalance} is the dissipated power. A system is passive if $\nabla^T_{\mathbf{x}} \mathcal{H}\,\mathbf{R(x)}\,\nabla_{\mathbf{x}} \mathcal{H} > 0$.

\section{Causal PH Impedance}


Hamiltonian dynamics formalism holds $\bm{f_{int}}$ as a system input. However, as long as a description had $\bm{f_{int}}$ as an outcome of  $\bm{x}_{d}$, it upholds the impedance causality for \eqref{eq:desired_dynamics} with the same Cartesian space energy of \eqref{eq:ImpedanceHam}. Following such a description is present.

\subsection{N-DoF MSD causal model}

Let $\bm{\momenta} = \bm{\Lambda_d} \dot{\bm{x}}$, and $\bm{\momenta}_d = \bm{\Lambda_d} \dot{\bm{x}}_{d}$, the Hamiltonian of \eqref{eq:desired_dynamics} with explicit input is:

\begin{multline}
    \CausalHamiltonian = \frac{1}{2} (\CausalDMomenta)^T \bm{\Lambda_d}^{-1}(\CausalDMomenta)\,+ \\
    \frac{1}{2} (\CausalDeviation)^T \bm{K_d}(\CausalDeviation)
    \label{eq:CausalImpedanceHam}
\end{multline}

Then, the PH equations with state vector $\begin{bmatrix}\bm{x} & \bm{\momenta}\end{bmatrix}^T$ are:

\begin{align}
    \begin{bmatrix}
        \dot{\bm{x}} \\ \dot{\bm{\momenta}}
    \end{bmatrix}
    &=
    \begin{bmatrix}
        \bm{0}_{k \times k} & \bm{I}_k \\
        -\bm{I}_k & -\bm{D_d}
    \end{bmatrix}
    \begin{bmatrix}
        \nabla_{\bm{x}}\,\CausalHamiltonian \\
        \nabla_{\bm{\momenta}}\,\CausalHamiltonian
    \end{bmatrix}
    +
    \bm{G}
    \begin{bmatrix}
        \bm{\momenta}_d \\
        \dot{\bm{\momenta}}_d \\
        \bm{f_{int}}
    \end{bmatrix} \label{eq:CausalStateInput}
    \\
    \bm{y} &= \bm{G}^T
    \begin{bmatrix} \bm{0}_{k} \\ \nabla_{\bm{\momenta}}\CausalHamiltonian\end{bmatrix}
    \label{eq:CausalOutput}
\end{align}

\noindent where

\begin{equation}
    \bm{G} =
    \begin{bmatrix}
        \bm{\Lambda_d}^{-1} & \bm{0} & \bm{0} \\
        \bm{0} & \bm{I}_k & \bm{I}_k
    \end{bmatrix}
    \label{eq:CausalInputMatrix}
    \hspace{1pt}, 
\end{equation}

\noindent and the input vector is $\bm{u} = \begin{bmatrix}\bm{\momenta}_d & \dot{\bm{\momenta}}_d & \bm{f_{int}}\end{bmatrix}^T$. The partial derivatives are $\nabla_{\bm{\momenta}}\CausalHamiltonian = -\nabla_{\bm{\momenta}_d}\CausalHamiltonian = \bm{\Lambda_d}^{-1}(\CausalDMomenta)$, and $\nabla_{\bm{x}}\CausalHamiltonian = -\nabla_{\bm{x}_{d}}\CausalHamiltonian = \bm{K_d}(\CausalDeviation)$. According to \eqref{eq:EnergyBalance}, the power balance for \eqref{eq:CausalStateInput} and \eqref{eq:CausalOutput} is:

\begin{align}
    \bm{y}^T \bm{u} &= \frac{d \CausalHamiltonian}{dt} + \nabla^T_{\bm{\momenta}}\CausalHamiltonian\,\bm{D_d}\,\nabla_{\bm{\momenta}}\CausalHamiltonian \nonumber \\
    (\dot{\bm{x}} - \dot{\bm{x}}_{d})^T\,(\dot{\bm{\momenta}}_d + \bm{f_{int}})
    &= \frac{d \CausalHamiltonian}{dt} +
    (\dot{\bm{x}} - \dot{\bm{x}}_{d})^T\bm{D_d}(\dot{\bm{x}} - \dot{\bm{x}}_{d})
    \label{eq:CausalBalance}
\end{align}


The rightmost term is a quadratic form with $\bm{D_d} \succ 0$, then it is positive $ \forall t$. The damping results in a dissipative system with respect to the interaction wrench and the inertial effect of the desired acceleration, as seen on the left side of \eqref{eq:CausalBalance}.

\subsection{Exogenous \textit{n}-DoF Passivity}

Regardless of the parity of \eqref{eq:CausalBalance}, in the presence of sensor errors, model uncertainty, and hardware constraints, relating stored energy to accumulated power exchange is more suitable, since the integral form of \eqref{eq:CausalBalance} avoids computing Cartesian accelerations, which depend on encoder derivatives.

Damping renders the dynamics passive with respect to the inputs, i.e., the exogenous power associated with $\bm{u}$. However, the desired pose $\bm{x}_d$ injects potential energy into the system without being on the input vector. Then, an integral, damping-independent expression for passivity can be:

\begin{equation}
    \int_0^t (\dot{\bm{x}} - \dot{\bm{x}}_{d})^T (\dot{\bm{\momenta}}_d + \bm{f_{int}})\,dt
    > \CausalHamiltonian(t)
    \label{eq:CausalPassivity}
\end{equation}

\subsection{Input-output power interconnection}

The equivalent PH description for the robot \eqref{eq:CanonicalHamiltonian} has power supply defined by:
\begin{equation}
    \left[\bm{y}^T \bm{u} \right]_{q} =  \dot{\bm{q}}^T\,\bm{\tau} \label{eq:qpower}
\end{equation}

The input $\bm{\tau}$ is the sum of the actuators input and interaction mapped torques. Then:

\begin{equation}
    \left[\bm{y}^T \bm{u} \right]_{q} =  \dot{\bm{q}}^T( \bm{J}^{T}\,\bm{f_{int}} + \bm{\tau}_{act}) = \dot{\bm{x}}^T\bm{f_{int}} + \dot{\bm{q}}^T\bm{\tau}_{act} \label{eq:qpower_with_fint}
\end{equation}

Combining \eqref{eq:qpower_with_fint} with the left-side of \eqref{eq:CausalBalance} one can see:

\begin{equation}
    \dot{\bm{q}}^T\bm{\tau}_{act} + \dot{\bm{x}}_{d}^T\bm{f_{int}} = \left[\bm{y}^T \bm{u} \right]_{q} - \left[\bm{y}^T \bm{u} \right]_{\Omega} - (\dot{\bm{x}}_{d} - \dot{\bm{x}})^T\dot{\bm{\momenta}}_d
    \label{eq:power_distribuition_full}
\end{equation}

Equation \eqref{eq:power_distribuition_full} shows how the supplied power, on the left-hand side, is distributed as follows: the first term on the right-hand side decouples the joint-space dynamics; the second term supplies the Cartesian impedance model input; and the third term accounts for the inertial effect of the desired Cartesian acceleration.


\section{Quasi-static reference analysis}


Equations \eqref{eq:CausalPassivity} and \eqref{eq:power_distribuition_full} are valuable for characterizing the robot \eqref{eq:EulerLagrange} controlled by \eqref{eq:cartesian_impedance_law}, for all $\bm{x}_{d}$, $\dot{\bm{x}}_{d}$, and $\ddot{\bm{x}}_{d}$. However, their generalization comes at the cost of increased complexity. In practice, the desired velocity and acceleration are often neglected, and the reference is defined solely by $\bm{x}_{d}(t)$. Therefore, in this section the equations are simplified for $\dot{\bm{x}}_{d}(t) = 0$ and $\ddot{\bm{x}}_{d}(t) = 0\,, \hspace{1pt} \forall t > 0$.

\subsection{Quasi-static passivity}

The cartesian impedance power supply \eqref{eq:CausalBalance} reduces to:

\begin{equation}
    \left[\bm{y}^T \bm{u} \right]_{\Omega} = \dot{\bm{x}}^T\,\bm{f_{int}}
    \label{eq:zpower_static}
\end{equation}

Then, from \eqref{eq:qpower_with_fint} it follows:

\begin{equation}
    \left[\bm{y}^T \bm{u} \right]_{\Omega} = \left[\bm{y}^T \bm{u} \right]_{q} - \dot{\bm{q}}^T\bm{\tau}_{act}
    \label{eq:power_distribuition_static}
\end{equation}

By definition, the integral form of \eqref{eq:power_distribuition_static} is:

\begin{equation}
    \CausalHamiltonian(t) + L_{\Omega}(t) = \JointHamiltonian(t) + L_{q}(t)
    -\int_0^t \dot{\bm{q}}^T\bm{\tau}_{act}\,dt
    \label{eq:energy_balance_static_1}
\end{equation}

\noindent where $L_{\Omega}(t)$ and $L_{q}(t)$ are the energy loss, by the cartesian impedance, and the robot Coriolis effects, respectively. Rearranging this energy balance and grouping together the loss one can see:

\begin{equation}
    \int_0^t \dot{\bm{q}}^T\bm{\tau}_{act}\,dt = \JointHamiltonian(t) - \CausalHamiltonian(t) + L(t)
    \label{eq:energy_balance_static}
\end{equation}

\noindent with $L(t) = L_{q}(t) - L_{\Omega}(t)$. Then, it holds the following:

\textit{Proposition 1}: System \eqref{eq:nDoFHam} connected to \eqref{eq:CausalImpedanceHam} by the control law \eqref{eq:cartesian_impedance_law} is passive with respect to the commanded power $\dot{\bm{q}}^T\bm{\tau}_{act}$ if

\begin{equation}
    \int_0^t \dot{\bm{q}}^T\bm{\tau}_{act}\,dt > \JointHamiltonian(t) - \CausalHamiltonian(t)
    \label{eq:CausalPassivity_static}
\end{equation}

\textit{Proof}: Let $L(t) > 0\,, \forall\,t > 0$, then

\begin{equation}
    \JointHamiltonian(t) - \CausalHamiltonian(t) + L(t) > \JointHamiltonian(t) - \CausalHamiltonian(t)
    \label{eq:CausalPassivity_static_proof}
\end{equation}

The left-hand side is defined by \eqref{eq:energy_balance_static}. The condition for $L(t)$ implies that the system is passive. Then, holding \eqref{eq:CausalPassivity_static} implies that the system is passive. \hfill $\blacksquare$

Differently from \eqref{eq:CausalPassivity}, this last passivity condition does not depend of the force-torque sensor data, neither an estimation of the interaction wrench using the impedance model.

\subsection{Quasi-static step response power}

Free motion is a suitable scenario for characterizing the power flow of the impedance controller. Although constrained interaction is also relevant, forces and torques do not necessarily perform mechanical work when applied perpendicular to the displacements. Considering \eqref{eq:desired_dynamics} under the quasi-static assumptions, and in free motion, the answer to the question \textit{what power is required to move the desired mass, freely, when a single-axis step for $\bm{x}_d$ is applied?} is achievable. Let the step input be on the $x$-axis, the response is

\begin{equation}
    x(t) = 1 - \frac{e^{- \zeta\,\omega_{n}\,t}}{\sqrt{1 - \zeta^2}}\,\cos{(\omega_d\,t - \beta)}
    \label{eq:step_response}
\end{equation}

\noindent in which $\omega^2_n = \frac{k_d}{m_d}$ is the system natural frequency, $\zeta = d_d\,\big( 2\,\sqrt{k_d\,m_d} \big)^{-1}$ is the damping ratio, $\omega_d = \omega_n \sqrt{1 - \zeta^2}$ is the damped natural frequency, and $\tan \beta = \zeta\, \big( \sqrt{1 - \zeta^2} \big)^{-1}$. At this conditions, the force acting on the desired mass is


\begin{equation}
    m_d\,\ddot{x} = -d_d\,\dot{x}(t) + k_d\,(x_d - x(t))
\end{equation}

Then, the step input entail the mechanical power:

\begin{equation}
    \mathcal{P}_{step}(t^*) = \dot{x}(t)\,\big[ k_d\,\big(x_d - x(t) \big) - d_d\,\dot{x}(t) \big]
    \label{eq:step_power}
\end{equation}

\noindent where $t^*$ is time after the step, and $x(t)$ and $\dot{x}(t)$ are according to \eqref{eq:step_response}. For notational convenience, the $x$ direction is chosen here, but \eqref{eq:step_power} is valid for any Cartesian coordinate, such as $y$, $z$, or angular components. Theoretically, an impedance-controlled robot subjected to a single-dimensional step in $\bm{x}_d$ exchanges this power without interaction. A model-based step power error can then be defined as:

\begin{equation}
    e_{step} = \mathcal{P}_{step}(t^*) - \dot{\bm{x}}^T \Big[ \bm{K_d}\,\big(\bm{x}_d(t^*) - \bm{x} \big) - \bm{D_d}\,\dot{\bm{x}}\Big]
    \label{eq:step_power_equivalence}
\end{equation}

Although the rightmost term of \eqref{eq:step_power_equivalence} accounts for the n-DoF power exchange ($\mathcal{P}_{\bm{x}}$), the dynamic decoupling of the diagonal Cartesian parameters allows the assumption that, with zero error in all directions except the one where the step is applied, the Cartesian power is the single-axis step power. In this way, the quasi-static step power analysis evaluates not only the rendered impedance parameters but also the joint-space dynamic decoupling, the accuracy of the robot model, and the execution of the control software, since all these factors affect the power flow for storage and dissipation when implementing \eqref{eq:cartesian_impedance_law}.

\section{Simulation validation}

We assess the applicability of \eqref{eq:CausalPassivity_static} and \eqref{eq:step_power_equivalence} in three scenarios: a robotic arm under a single-step reference to show the step power fidelity evaluation, a quadruped leg following a periodic foot trajectory, and the same leg under a step up-and-down input sequence that induces a jump of nearly one meter. Passivity is evaluated in all cases. For broad adoption of the metrics, simulations were conducted in Gazebo with ROS2 \cite{ROS2, ros2control, pinocchioweb}. The impedance controller was implemented in C++\footnote{Available on: github.com/qleonardolp/ros2\_impedance\_controller}, with a control loop period of \SI{1}{\milli\second}. Simulation datasets are available online \footnote{Simulation datasets: github.com/qleonardolp/ICAR2025\_PHBenchmark}.

\subsection{Robotic arm: single-step}

\begin{figure}[t]
    \centering
    \includegraphics[width=\linewidth]{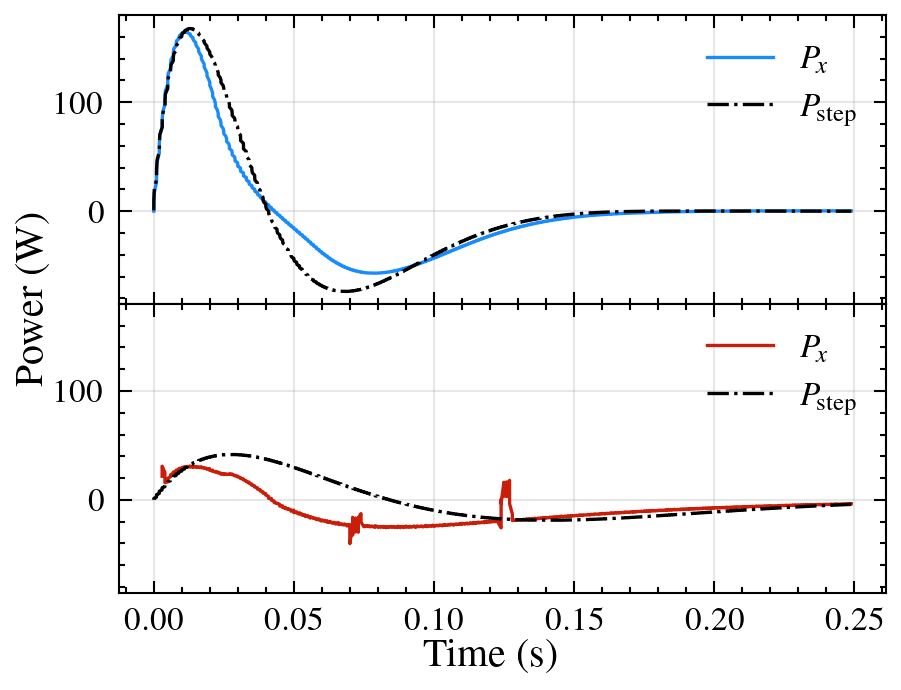}
    \caption{Step power characterization of a 6-DoF robotic arm with (top) and without (bottom) inertia shaping (IS). Dashed lines are the step reference power. With IS the root mean square error of \eqref{eq:step_power_equivalence} was\,\SI{13.114}{\watt}, and without IS was\,\SI{18.016}{\watt}, both over the \SI{0.25}{\second} interval shown. Tuning to achieve a stable step response without IS was more difficult than with IS. Then, the Cartesian power without IS displayed some spikes (solid red line).}
    \label{fig:sim_step_power}
\end{figure}

The 6-DoF robotic arm starts at rest with zero deviation in Cartesian impedance, i.e., $\CausalHamiltonian(0) = 0$. A single-axis step input is then applied, with an amplitude of \SI{0.4}{\meter}. Fig. \ref{fig:sim_step_power} shows the step signal, the step reference power \eqref{eq:step_power}, and the controlled Cartesian power after the step with and without IS. Impedance parameters are in Tab. \ref{tab:robotic_arm_params} and \ref{tab:robotic_arm_params_without_is}. With IS, the desired mass is approximately half the mass of the robot, yet the fidelity was higher. On the other hand, the absence of IS implies a cross-axis power exchange due to $\bm{\Lambda}(\bm{x})$ coupling the impedance. Thus, step power fidelity is less suitable without inertia shaping.

\begin{table}[hb]
\centering
\caption{Robotic arm impedance parameters with IS.}
\label{tab:robotic_arm_params}
\begin{tabular}{lll}
\hline
\textbf{} &
\multicolumn{1}{c}{\textbf{Translational ($x$ $y$ $z$)}} &
\multicolumn{1}{c}{\textbf{Rotational ($\varphi$ $\theta$ $\psi$)}}\\ \hline
$\bm{K_d}$  & 800.0 800.0 800.0 [\SI{}{\newton\per\meter}] & 120.0 120.0 120.0 [\SI{}{\newton\per\rad}] \\
$\bm{D_d}$  & 134.2 134.2 134.2 [\SI{}{\newton\second\per\meter}] & 13.96 13.96 13.96 [\SI{}{\newton\second\per\rad}] \\
$\Lambda_d$ & 10.00 10.00 10.00 [\SI{}{\kilogram}] & 0.722 0.722 0.722 [\SI{}{\kilogram\meter\squared}] \\ \hline
\end{tabular}
\end{table}

\begin{table}[ht]
\centering
\caption{Robotic arm impedance parameters without IS.}
\label{tab:robotic_arm_params_without_is}
\begin{tabular}{lll}
\hline
\textbf{} &
\multicolumn{1}{c}{\textbf{Translational ($x$ $y$ $z$)}} &
\multicolumn{1}{c}{\textbf{Rotational ($\varphi$ $\theta$ $\psi$)}}\\ \hline
$\bm{K_d}$  & 400.0 400.0 400.0 [\SI{}{\newton\per\meter}] & 70.0 70.0 40.0 [\SI{}{\newton\per\rad}] \\
$\bm{D_d}$  & 134.2 134.2 134.2 [\SI{}{\newton\second\per\meter}] & 15.08 15.08 15.08 [\SI{}{\newton\second\per\rad}] \\
$\Lambda_d$ & \multicolumn{1}{c}{$\Lambda(\bm{x})$} & \multicolumn{1}{c}{$\Lambda(\bm{x})$} \\ \hline
\end{tabular}
\end{table}

\subsection{Quadruped leg: step trajectory}

\begin{figure}[b]
    \centering
    \includegraphics[width=\linewidth]{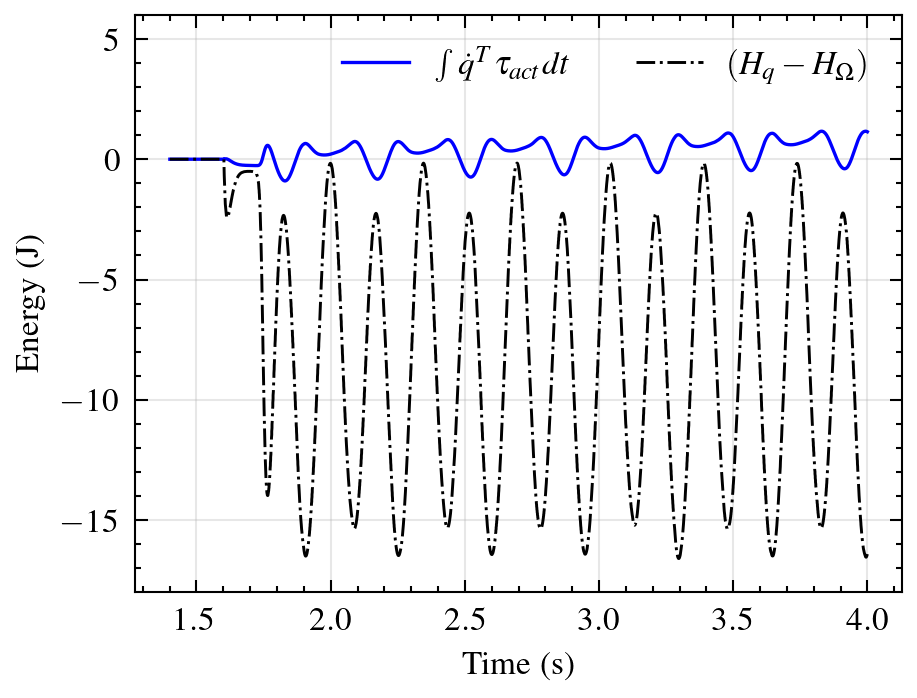}
    \caption{Passivity characterization of the 3-DoF quadruped leg following a CPG trajectory. The integrated command power (solid blue) is non-monotonic but shows an increasing average over time, while the Hamiltonian gap (dot-dashed black) confirms the controller’s growing passivity.}
    \label{fig:gait_passivity}
\end{figure}

In this scenario, the quadruped leg (3-DoF) is suspended and the foot is free to move. The foot trajectory follows the Central Pattern Generator (CPG) model \cite{zhang2024online}. In this case, joint control does not require solving inverse kinematics; instead, joint torques are applied according to the impedance between the leg support and the desired foot frame. The impedance parameters are listed in Tab. \ref{tab:quadruped_leg_params}. Step length is \SI{0.40}{\meter} and step period is \SI{0.7}{\second}. The integrated command power and the Hamiltonian function gap are shown in Fig. \ref{fig:gait_passivity}. Although the integrated command power is non-monotonic, its average increases over time. In other words, the controller, already passive from the beginning, gradually becomes more passive in accordance with \eqref{eq:CausalPassivity_static}, even under a periodic reference.

\begin{table}[t]
\centering
\caption{Quadruped leg impedance parameters (without IS).}
\label{tab:quadruped_leg_params}
\begin{tabular}{ll}
\hline
\textbf{} &
\multicolumn{1}{c}{\textbf{Translational ($x$ $y$ $z$)}} \\ \hline
$\bm{K_d}$  & 400.0 400.0 800.0 [\SI{}{\newton\per\meter}] \\
$\bm{D_d}$  & 43.00 43.00 90.00 [\SI{}{\newton\second\per\meter}] \\ \hline
\end{tabular}
\end{table}

\subsection{Quadruped leg: jumping}

\begin{figure}[t]
    \centering
    \includegraphics[width=0.83\linewidth]{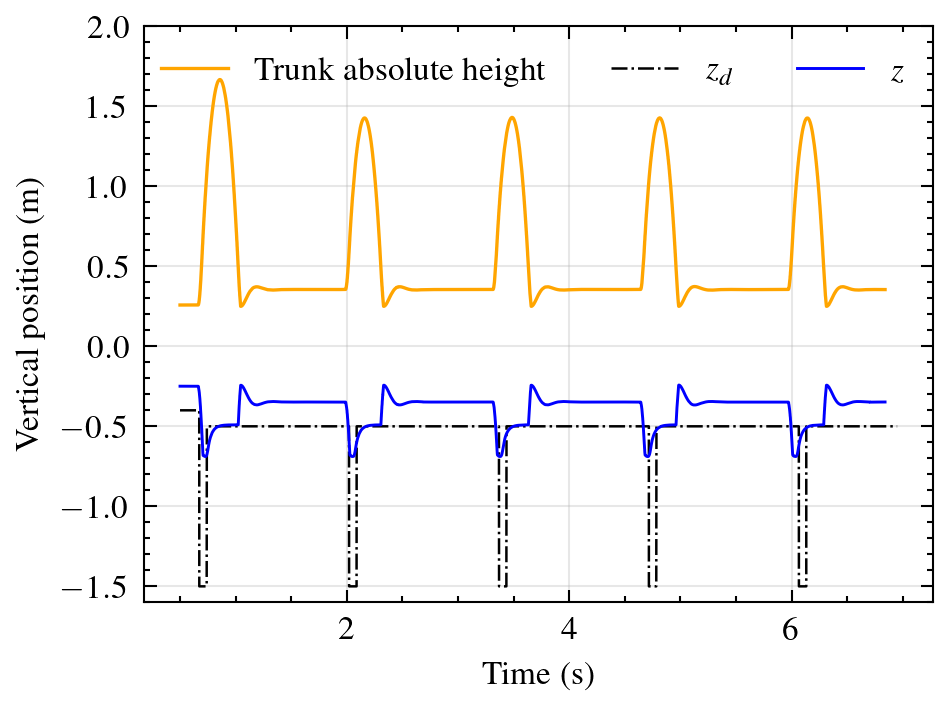}
    \caption{Jumping demonstration of the 3-DoF quadruped leg. Foot vertical reference (dashed black) and state (solid blue) are relative to the robot trunk, or the leg support in this case. In solid orange is the trunk absolute height.}
    \label{fig:jumping_position}
\end{figure}

\begin{figure}[H]
    \centering
    \includegraphics[width=\linewidth]{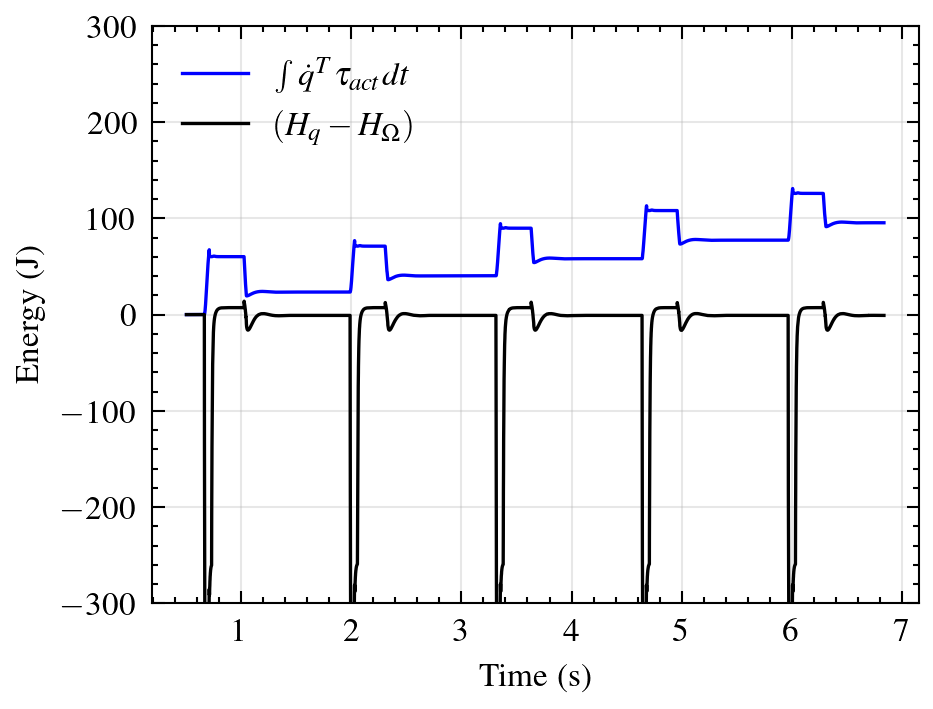}
    \caption{Passivity characterization of the 3-DoF quadruped leg performing jumps. The integrated command power (blue) increases over time. The Hamiltonian gap (black) remains below the integrated command power, satisfying the passivity condition, and the valleys show the jump demand.}
    \label{fig:jumping_passivity}
\end{figure}

The jumping demonstration used the same leg model and impedance parameters (Tab. \ref{tab:quadruped_leg_params}). Considering the trunk-relative position of the foot frame, a step-down input of \SI{1}{\meter} was applied, followed by a step-up \SI{0.2}{\second} later to return the foot to its resting relative position. In this way, upon ground contact, the vertical impedance spring was relaxed and the leg was in a suitable configuration for landing. The foot vertical reference and state, as well as the leg support (trunk) absolute height, are shown in Fig. \ref{fig:jumping_position}. The foot vertical position reaches the resting relative position of \SI{-0.5}{\meter} a few tenths of a second before ground contact. When jumping, the impedance Hamiltonian $\CausalHamiltonian$ peaks at \SI{536.62}{\joule}. Since the negative of $\CausalHamiltonian$ appears in the Hamiltonian gap, these energy peaks are represented as valleys, as shown in Fig. \ref{fig:jumping_passivity}. Once again, the integrated command power increases gradually, displaying passivity.

Despite the computational load of the Hamiltonian gap, the controller runtime diagnostics did not miss the \SI{1}{\milli\second} loop period during the simulations, even without IS, which demands the computation of the task space inertia matrix $\bm{\Lambda}_{d} = \bm{\Lambda}(\bm{x}) = \bm{J}(\bm{q})^{-T}\,\bm{M}(\bm{q})\,\bm{J}(\bm{q})^{-1}$. Moreover, since the metrics are model-based and online (state-dependent), both passivity and fidelity metrics are sensitive to state noise and model accuracy.

\section{Conclusion and future work}

This work proposed a physics-centered analysis for benchmarking impedance control based on the Port-Hamiltonian formalism. A causality-consistent PH model was introduced to describe mass-spring-damper impedance at Cartesian coordinates, from which two metric candidates were derived: a differentiable n-DoF passivity condition and a free motion fidelity. The passivity condition accounts for time-varying references and does not rely on force-torque sensing, while the fidelity metric captures the accuracy of the impedance response through step-power evaluation. Both were demonstrated in simulations with a robotic arm and a quadruped leg, covering scenarios of free motion, for step and trajectory reference, and impacts. Results expressed the suitability of the PH framework for standardized, application-agnostic benchmarking of impedance control. Future work includes a comparative study against classical metrics, experimental analysis using physical platforms, and exploring the integration of these systems with learning-based controllers and energy storage tanks.




\section*{Acknowledgment}

The authors would like to thank the Legged Robotics Group of the Robotics Laboratory of the São Carlos School of Engineering, University of São Paulo.

\typeout{}
\bibliographystyle{bib/IEEEtran}
\bibliography{bib/references}

\end{document}